\documentclass[10pt, a4paper]{article}

\usepackage[]{lrec2026}
\usepackage{multirow}
\usepackage{booktabs}
\usepackage[inline]{enumitem}
\usepackage{xspace}
\usepackage{twemojis}
\usepackage{caption}
\usepackage{subcaption}

\newcommand{\datasetname}{\textsc{SciLaD}\xspace}
\newcommand{\generalModelName}{\textsc{SciLaD-M}\xspace}
\newcommand{\modelNameCustomTokenizer}{\textsc{SciLaD-M (custom)}\xspace}
\newcommand{\modelNameRobertaTokenizer}{\textsc{SciLaD-M (RoBERTa)}\xspace}

\newcommand{\cleanPlainDataset}{plain text clean split\xspace}

\title{\datasetname{}: A Large-Scale, Transparent, Reproducible Dataset\\ for Natural Scientific Language Processing}

\name{
  Luca Foppiano\textsuperscript{3,1} \quad Sotaro Takeshita\textsuperscript{5} \quad Pedro Ortiz Suarez\textsuperscript{4}\\
  \large \textbf{Ekaterina Borisova\textsuperscript{1} \quad Raia Abu Ahmad\textsuperscript{1} \quad \textbf{Malte Ostendorff}\textsuperscript{4}}\\
  \large \textbf{Fabio Barth\textsuperscript{1} \quad Julian Moreno-Schneider\textsuperscript{1} \quad Georg Rehm\textsuperscript{1,2}}\\
}

\address{\textsuperscript{1}Deutsches Forschungszentrum für Künstliche Intelligen GmbH (DFKI), Germany \\
\textsuperscript{2}Humboldt-Universität zu Berlin, Germany ~~~ \textsuperscript{3}ScienciaLAB, Portugal \\
\textsuperscript{4}Common Crawl Foundation, USA ~~~
\textsuperscript{5}University of Mannheim, Germany \\
} 

\abstract{\datasetname{} is a novel, large-scale dataset of scientific language constructed entirely using open-source frameworks and publicly available data sources. It comprises a curated English split containing over 10 million scientific publications and a multilingual, unfiltered TEI XML split including more than 35 million publications. We also publish the extensible pipeline for generating \datasetname{}. The dataset construction and processing workflow demonstrates how open-source tools can enable large-scale, scientific data curation while maintaining high data quality. Finally, we pre-train a RoBERTa model on our dataset and evaluate it across a comprehensive set of benchmarks, achieving performance comparable to other scientific language models of similar size, validating the quality and utility of \datasetname{}. We publish the dataset and evaluation pipeline to promote reproducibility, transparency, and further research in natural scientific language processing and understanding, including scholarly document processing.\\ 
\newline
\Keywords{Dataset, scientific, natural scientific language processing, LLM, encoder, text mining}}

\begin{document}

\maketitleabstract

\section{Introduction}
High-quality accessible datasets are a cornerstone of progress in natural language processing (NLP). In recent years, the availability of large-scale open corpora has enabled significant advances across tasks such as machine translation, summarization, and question answering \cite{raffel2023exploringlimitstransferlearning,devlin-etal-2019-bert}. However, while much of this progress has been driven by general-domain data, scientific language remains comparatively underrepresented in open datasets, limiting the reproducibility and transparency of research in specialized domains \cite{lo2020s2orc,beltagy2019scibert}.

Scientific text exhibits unique linguistic and structural properties -- precise terminology, formal discourse organization, and data-centered argumentation -- that differ markedly from everyday language. Creating large, openly available, and well-structured corpora of scientific publications is, thus, essential for training and evaluating language models in the scientific domain \cite{fisas-etal-2015-discoursive}. Resources such as S2ORC \cite{lo2020s2orc} and CORE \cite{knoth2023core} have demonstrated the value of openly licensed scientific data; however, few offer fully transparent, reproducible pipelines built exclusively on open-source frameworks.

We present \datasetname{} (Scientific Language Dataset), a large-scale dataset of scientific publications created entirely using open-source tools, ensuring reproducibility, transparency, and accessibility. The dataset is harvested from Open Access (OA) scientific articles and represented in  TEI XML (Text Encoding Initiative), preserving document structure and metadata, which is essential for downstream applications such as citation graph analysis, bibliometric research, and retrieval-based tasks. 


To demonstrate the utility and quality of \datasetname{}, we train and evaluate a RoBERTa-base \cite{liu2019roberta} encoder model. The model and its tokenizer are pre-trained from scratch using our data, \modelNameCustomTokenizer. For comparison, we also train a variation, which uses the original RoBERTa tokenizer, \modelNameRobertaTokenizer.
The model \modelNameCustomTokenizer achieves results comparable to other domain-specific encoders such as SciBERT \cite{beltagy2019scibert}, validating the effectiveness of the dataset as a foundation for scientific NLP. Furthermore, we introduce a reproducible evaluation pipeline to benchmark encoder and encoder-decoder architectures on multiple scientific tasks, providing a framework for systematic comparison of models trained on scientific corpora.

Our main contributions are:

\begin{itemize}
    \item A large-scale multilingual corpus of over 35 million scientific articles sourced from Unpaywall\footnote{\url{https://unpaywall.org}}, structured in TEI XML format\footnote{\url{https://tei-c.org}}, suitable for various upstream and downstream tasks, including bibliographic, citation graph analysis (109 million citations), and citation graph or characterization training, as well as classic NLP tasks in the scientific domain. We also created a filtered English split of 10 million text documents for pre-training.
    \item An open source pre-processing pipeline for the dataset using frameworks such as Grobid \cite{Grobid} and Datatrove \cite{penedodatatrove} as well as an evaluation pipeline to fine-tune and evaluate encoder- or encoder-decoder-based models on common scientific NLP tasks.
    \item Encoder-based scientific language models based on RoBERTa-base architecture \cite{liu2019roberta} (110M parameters) and pre-trained from scratch with a clean, plain text dataset of over 10 million scientific publications in English. We evaluate the produced models on multiple benchmarks and compare them with a set of baseline models. 
\end{itemize}


The resources shared through this article are described in Section~\ref{sec:availability} and will be made publicly available upon acceptance. 

\section{Dataset Construction}

The dataset was constructed using the open-source \datasetname{} pipeline, which automates the retrieval, filtering, and preprocessing of scientific publications. The pipeline integrates multiple components for metadata extraction, text cleaning, and language identification, ensuring consistent data quality across languages and scientific domains. By relying exclusively on open-access sources and transparent workflows, we enable reproducibility and facilitate community-driven extensions of the dataset. In the following, we provide brief descriptions of the pipeline architecture and each stage of the data construction process, from harvesting to preprocessing and evaluation.

\begin{figure*}[th]
  \begin{center}
   \includegraphics[width=1\textwidth]{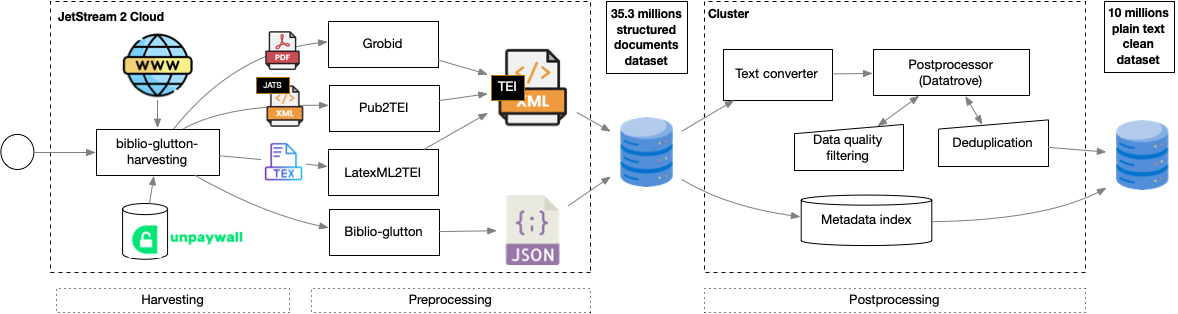}
  \end{center}
  \caption{The \datasetname{} pipeline. The data is harvested by accessing direct links to the source files and is transformed into TEI XML. The structured format's scientific data is streamed to text, while its bibliographic data are aggregated in a metadata index. Finally, the text is filtered and deduplicated.}
  \label{fig:data-collection-and-processing}
\end{figure*}

\subsection{Processing pipeline}
\label{subsec:proc_pipe}

The pipeline is composed of three steps (Figure~\ref{fig:data-collection-and-processing}). First, articles are collected and downloaded following the Unpaywall snapshot and complemented with additional formats, including JATS (Journal Article Tag Suite) XML and \LaTeX{} from other sources, e.\,g., arXiv and PubMed Central (PMC). Unpaywall is a database that indexes over 50 million open-access scholarly articles by aggregating metadata from trusted repositories, journals, and archives. It provides DOI-based access to legally available full texts, which we used as the primary data source for collecting multilingual documents across scientific domains. In the second step, these articles were then transformed into a standardized, structured format (TEI XML). In the third step, the data was converted into plain text. After extracting the plain text, we applied data quality filtering and deduplication.

\subsubsection{Harvesting}
\label{subsec:harvesting}

The main challenge in accessing scientific publications is that they are typically published behind paywalls, restricting automated access. However, a significant number of publications are now available in OA \cite{piwowar2018state}, making it possible to access these resources legally while complying with constraints related to text and data mining and fair use. However, even for OA publications, the dominant scientific publishers often use mechanisms to limit access (access limitation and strict policies), especially via programmatic means, which implies the need for robust harvesting methods. 

As illustrated in Figure \ref{fig:data-collection-and-processing}, we rely on Unpaywall for accessing PDF files \cite{piwowar2018state, else2018unpaywall, chawla2017unpaywall}. We used the full snapshot produced on 14 June 2023, a total of 38.9 million entries with a non-empty \texttt{url\_for\_pdf} field. These entries, however, are not only documents but also other ``components'' that represent elements of documents (figures, tables, etc.), see Table~\ref{tab:unpaywall-stats}. After removing these entries and downloading the PDF files corresponding to each URL, we obtained 31.6 million documents. These documents are, at this stage, neither filtered nor deduplicated. Although the majority (77.1\%) of the data is in English, there are still some texts in other languages (see Figure~\ref{sub:fig:language-distribution}). To facilitate access to \LaTeX{} and JATS XML we mirrored arXiv\footnote{\url{https://info.arxiv.org/help/bulk_data_s3.html}} resources (around 2 million OA articles with \LaTeX{} sources), PMC\footnote{\url{http://pmc.ncbi.nlm.nih.gov}} full texts (around 4 million articles in JATS XML) and the PLOS\footnote{\url{https://github.com/PLOS/allofplos}} \cite{elizabeth_seiver-proc-scipy-2018} XML full text collections (around 300,000 articles), all available publicly online, on different AWS (Amazon Web Service) S3 buckets. 

\begin{table}[ht!]  
\centering
\resizebox{\columnwidth}{!}{
\begin{tabular}{lc}
\toprule
\textbf{Description} &  \textbf{Count} \\ 
\midrule
Total Unpaywall entries (with non-empty \texttt{url\_for\_pdf}) & 38,931,157  \\ 
Total entries (excluding type ``component'')         & 37,451,475 \\
Total harvested PDF documents                        & 31,653,759 \\
\bottomrule
\end{tabular}}
\caption{Overview of the PDF files collected from the Unpaywall snapshot. The entries with a non-empty URL-to-PDF are retained, then filtered to remove ``components'' and further reduced by about 6 million documents that fail to download due to incomplete URLs, limiting landing pages, etc.}
\label{tab:unpaywall-stats}
\end{table}

The harvesting and processing of these files has been performed on the JetStream 2 HPC environment. We divided the Unpaywall snapshot into 3,000 partitions and used 24 virtual machine instances in parallel to harvest and process the set of documents. Each instance had 8 CPUs, 32 MB of RAM, and one GPU with 8 GB of VRAM. 

\subsubsection{Preprocessing}
\label{subsec:preprocessing}

The biggest challenge in extracting a structured representation of published content is the immense variety of formats and presentation layers in scientific publications, which makes it very difficult to reliably extract and normalize content into a single structured format. The extraction of content from arbitrary PDF files is a well-known challenge \cite{Westergaard_2017}. As mentioned, we normalize the three heterogeneous data types (PDFs, JATS XML, and \LaTeX{}) to a standardized, structured format, i.\,e., TEI XML, which we then transform into a text format in the preprocessing step.

Given the extreme volume of publications to process, we did not use Visual Language Models (VLM) to extract the text. Using VLMs may yield higher accuracy compared to other LLM-based approaches, however, the processing is too slow and too expensive to scale beyond 500 million pages \cite{blecher2023nougat}. For example, Nougat, a Visual Transformer that focuses on scientific papers, requires an A10 with 24GB VRAM to process six pages per batch, with a processing time of 19.5 seconds per batch, i.\,e., 3.25 seconds per page \cite{blecher2023nougat}. Other API-based LLMs, such as ChatGPT, also offer the possibility of extracting text from PDFs, but their closed-source nature makes the reproducibility of our work impossible, and the costs for extraction are much too high. 


We, therefore, decided to use an open source software that is not based on a VLM: Grobid \cite{Grobid} relies on a set of custom Recurrent Neural Network (RNN) and Conditional Random Field (CRF) models that work well on commodity hardware. Grobid can process, on average, around 120 pages per second on a single commodity server, which is approximately 400 times faster than Nougat.  

It is considered that VLMs offer more reliable extraction of formulas and figures. However, \cite{olmocr} shows that, for end-to-end tasks, results with Grobid-extracted full texts (based on an old version of Grobid from 2019) are comparable to those of the recent state-of-the-art VLM olmOCR. As Grobid extracts the text layer of a PDF file, it avoids errors coming from the Optical Character Recognition (OCR) processing. Nowadays, the large majority of scientific PDF files are born digital, i.\,e., they have a reliable text layer. With Grobid \cite{Grobid}, we processed roughly 31 million scientific articles (Table~\ref{tab:tei_xml_stats}).

Although \LaTeX{} and JATS XML are already structured formats, we also transformed them into TEI XML. For converting JATS XML to TEI XML, we used Pub2TEI, a framework that standardizes heterogeneous XML representations from different scientific publishers into TEI XML \cite{pub2tei}. It relies on a comprehensive set of XSLT stylesheets that map publisher-specific metadata and structural elements -- such as bibliographic records, abstracts, citations, and full texts -- into a consistent schema. For the \LaTeX{} to TEI XML conversion, we used \LaTeX{ML}, a software system designed to convert \LaTeX{} documents into structured XML representations suitable for further processing and web publication \cite{ginev2014ebooksgraphicslatexml}. It emulates the behavior of the \TeX{} engine to interpret document content and structure, producing an intermediate XML format that preserves both textual and mathematical semantics. This format can then be transformed, among others, into TEI XML.

\begin{table}[ht]  
\centering
\resizebox{\columnwidth}{!}{
\begin{tabular}{lc}
\toprule
\textbf{Description} &  \textbf{Count} \\ 
\midrule
Grobid (PDF files from Unpaywall)      & 31,062,728 \\
Pub2TEI (JATS XML files from PMC/PLoS) & 3,008,154 \\
\LaTeX{ML} (\LaTeX{} files from arXiv) & 1,256,667 \\
\midrule
\textbf{Total} & 35,342,549 \\
\bottomrule
\end{tabular}}
\caption{Distribution of structured documents by source type. The Grobid row originates from ``Total harvested documents'' in Table~\ref{tab:unpaywall-stats} where about 500,000 documents have not been processed by Grobid, likely because they are corrupted, do not contain any, or too much text (> 1.5 M tokens), etc.}
\label{tab:tei_xml_stats}
\end{table}

\begin{figure*}[htbp]
    \centering
    \subfloat[Language distribution]{\includegraphics[width=0.45\textwidth]{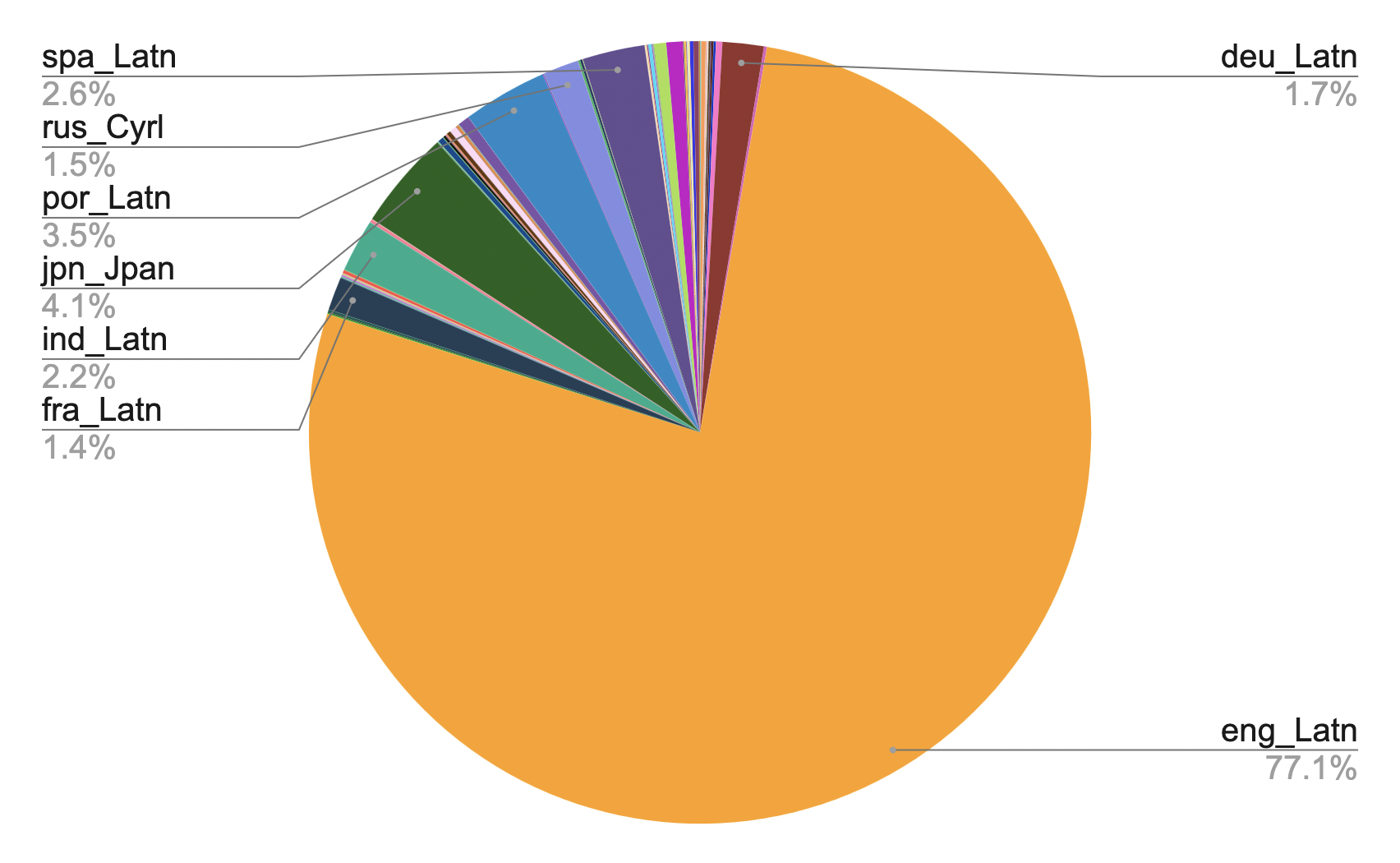}\label{sub:fig:language-distribution}}
    \subfloat[License distribution]{\includegraphics[width=0.45\textwidth]{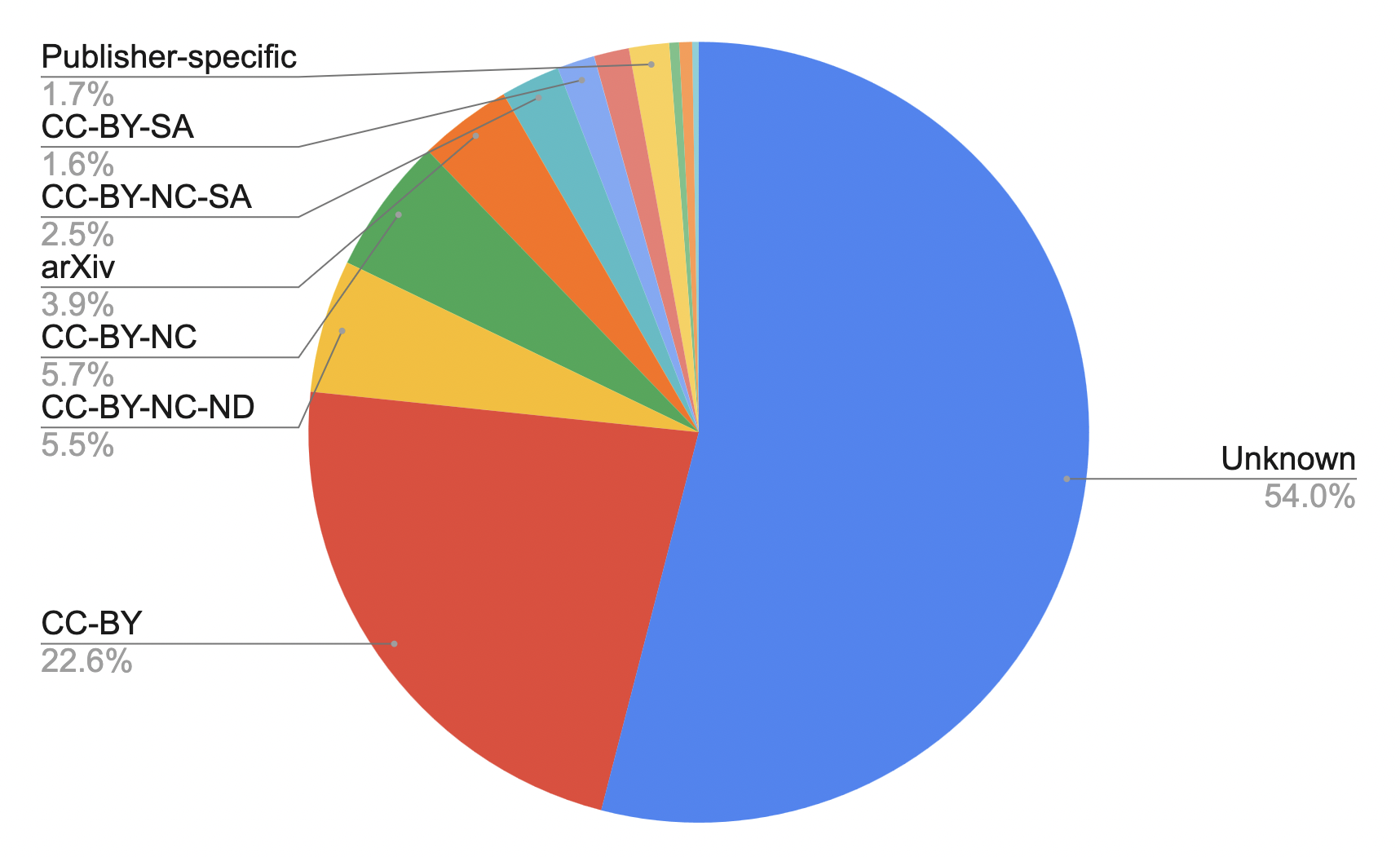}\label{sub:fig:licence-distribution}} \\
    \caption{Distribution of licenses and languages of the harvested dataset.}
    \label{fig:harvested-dataset-statistics}
\end{figure*}

After harvesting and preprocessing the dataset, we obtained 35.3 million publications in TEI XML (Table~\ref{tab:tei_xml_stats}). The dataset's original format was unequally distributed, with a predominance of PDF with 87\% (31 million), followed by JATS XML and \LaTeX{}, 8.5\% (3 million) and 3.6\% (1.2 million), respectively (see Table~\ref{tab:tei_xml_stats}). English was the most common language, followed by Japanese, Portuguese, and Spanish (Figure~\ref{sub:fig:language-distribution}). The TEI XML provides a structured format with 109 million nodes (references cited) and 1.51 billion edges linking citations with their contextualized sentences. Furthermore, we obtained approximately 19 million unique keywords assigned by authors and published in the headers of the articles.

Figure~\ref{sub:fig:licence-distribution} illustrates the distribution of licenses, with particular attention to OA licenses. Although Creative Commons (CC) licenses account for 37.9\%, the majority of articles have unidentified licenses (54\%). From our study, it also emerges that reusable content accounts for 32.4\% of the harvested OA documents. We want to point out that licensing is a gray area that is often overlooked; not all OA content can be redistributed, and the information is, in most cases, challenging to find because it is scattered around several places: articles, landing pages, and the respective publisher's or journal's policies on data sharing. At this stage the dataset consists of 35.3 million publications in TEI XML (Table~\ref{tab:tei_xml_stats}), including PDF documents and data transformed from JATS XML and \LaTeX{}.

\subsubsection{Postprocessing}
\label{subsec:postprocessing}

We prepared the harvested data for pre-training by converting the documents into plain text, selecting only English documents, applying text quality filtering, and deduplication steps. 

\paragraph{TEI XML to Text} The TEI XML to Text conversion was performed using a configurable streaming process to retain only minimal structure. During text processing, all paragraphs were collected, including captions, while formulas, equations, and table bodies were excluded. Equations and table content strongly depended on how the PDF file was created, making it challenging to assess proper quality extraction, avoiding a decrease of text quality. Within each paragraph, detected sentences were joined together, with a single space when necessary. References were maintained in the text through their original callout formats. During conversion, we computed weighted language detection scores aggregated from line-by-line results \cite{abadji2022towards} using an optimized version of fasttext \cite{joulin2016fasttext,burchell-etal-2023-open}.

\paragraph{Filtering} The filtering was divided into two parts. We first filtered all non-English text from the corpus and then implemented a pipeline with Datatrove \cite{penedodatatrove}, a scalable open-source framework for data cleaning and preparation. Initial filtering was performed using the language-detection scores. All documents with a language detection confidence score below 80\% were excluded. This allowed the removal of all documents whose output was deemed unreliable due to character encoding, encryption and incorrect OCR results performed in the past. Furthermore, we applied the Gopher language quality filtering \cite{rae2021scaling} with the same parameters as described in the original work. We did not use any other filters provided by Datatrove, such as C4 \cite{raffel2023exploringlimitstransferlearning} or FineWeb \cite{penedo2024finewebdatasetsdecantingweb}, because they are designed for web data, which is expected to be noisier than text from scientific articles. 


\paragraph{Deduplication} We applied MinHash deduplication on the document level to each individual filtered dump. We utilized the deduplication pipeline from Datatrove \cite{penedodatatrove}, which was also used for FineWeb \cite{penedo2024finewebdatasetsdecantingweb}. For this, we generated 5-grams using a word tokenizer \cite{bird2009natural} and computed MinHashes using 112 hash functions, the same setup was used for the FineWeb deduplication. We also targeted documents that are at least 75\% similar.

\subsection{Corpus Overview}

The final corpus obtained for English consisted of 68.7 billion tokens, comprising 10,999,210 documents (with an average of approximately 6,000 tokens per document). Filtering and deduplication reduced the corpus of about 20 million English articles by 50\%. 
The pipeline steps, including data harvesting and TEI XML to text conversion, are shared on GitHub, as discussed in Section~\ref{sec:availability}. 

\section{Experimental Setup}

We pre-trained an encoder-based model to showcase our dataset's utility for model development. To evaluate the pre-trained language model, we developed a comprehensive evaluation pipeline that is easy to use and publicly available. Unlike current benchmarking pipelines, such as Eval-Harness \cite{eval-harness}, our pipeline focuses on scientific benchmarks where models need to be fine-tuned for downstream task.

Our goal is to demonstrate that the corpus is a valuable asset for the research and open-source community, which is why we decided not to train a generative model but rather to use a model architecture comparable to established scientific language models \cite{beltagy2019scibert}. This decision was also supported by a lower economic footprint and reduced costs for reproducibility.

In this section, we first describe the pre-training process and then outline the evaluation pipeline and benchmarking process.

\subsection{Language Model Pre-Training} 

\paragraph{Language Models}

To assess the quality of our dataset, we pre-train the model from scratch (with random weights initialization) instead of starting from an existing checkpoint. We obtain \modelNameCustomTokenizer by training both the model and the tokenizer using our data. The resulting custom tokenizer has the same vocabulary size as the original RoBERTa tokenizer, of 50265 tokens. To make the comparison fairer and transparent, we also train a version that uses the original RoBERTa tokenizer: \modelNameRobertaTokenizer. 

We use the same model architecture and training objective as RoBERTa~\citep{liu2019roberta}, namely the \texttt{base} architecture\footnote{\url{https://huggingface.co/FacebookAI/roberta-base}}, composed of a stack of 12 Transformer layers with self-attention, trained on a masked language modeling (MLM) objective.

We follow the hyperparameters described in the original RoBERTa configuration, and we set the batch size ($=256$), with $2$ gradient accumulation steps, which makes the effective batch size ($=1024$). We trained the model on two H200 with 80Gb VRAM GPUs. We split our dataset into 80/10/10 (\%) for training, validation, and held-out test splits, respectively.



\subsection{Evaluation Pipeline}

We provide an open and user-friendly evaluation pipeline that allows researchers to fine-tune and evaluate language models on our dataset. The pipeline is built entirely using open-source components and supports a wide range of encoder, decoder, and sequence-to-sequence architectures (e.\,g., \texttt{BERT} \cite{devlin-etal-2019-bert}, \texttt{T5} \cite{raffel2023exploringlimitstransferlearning}, \texttt{mBART} \cite{liu2020multilingualdenoisingpretrainingneural}). It can be easily deployed online or locally, ensuring reproducibility and comparability across experiments.

The evaluation interface standardizes preprocessing, tokenization, and scoring across tasks, producing unified metrics that facilitate benchmarking multilingual scientific language understanding.

\paragraph{Benchmarks}

To evaluate the models, we followed the benchmarks used by \citet{beltagy2019scibert}, which focus on \begin{enumerate*} \item Named Entity Recognition (NER); \item Participants, Interventions, Comparisons, and Outcomes (PICO) Extraction; \item Dependency Parsing (DEP); \item Relation Classification (REL); and \item Text Classification (CLS).\end{enumerate*} 
We released an evaluation framework (see Section~\ref{sec:availability}) that fine-tunes Hugging Face models on all reported benchmarks, outputting their evaluation results.

NER is evaluated using four different benchmarks, three of which tackle the biomedical field, while the last is focused on computer science (CS). BC5CDR \cite{li2016biocreative} is a widely-used benchmark for evaluating biomedical NER and relation extraction (RE), consisting of 1,500 PMC abstracts manually annotated for chemical and disease mentions as well as their interactions. Similarly, JNLPBA \cite{collier2004introduction} evaluates biomedical NER, consisting of 2,000 MEDLINE abstracts annotated with five entity types: protein, DNA, RNA, cell type, and cell line. NCBI-disease \cite{dougan2014ncbi} focuses on disease entities in PMC abstracts, with 6,892 disease mentions mapped to 790 unique disease concepts. Finally, SciERC \cite{luan2018multi} is a benchmark for joint scientific extraction tasks, comprising 500 scientific abstracts from AI conference proceedings. It combines NER with RE and co-reference resolution, and includes six entity types such as Task, Method, and Metric, with seven relation types such as Compare and Part-of. 

PICO is a NER-like task that focuses on clinical trial papers. To evaluate it, we use the EBM-NLP benchmark \cite{nye2018corpus}, consisting of 5,000 annotated abstracts. For DEP evaluation, we use GENIA \cite{kim2003}, a semantically annotated corpus of 2000 abstracts from biological articles. REL is evaluated on two datasets, ChemProt \cite{Kringelum2016ChemProt30AG}, a benchmark with chemical-protein-disease annotations, and SciERC. Finally, CLS is evaluated using ACL-ARC \cite{jurgens2018measuring} and SciCite \cite{cohan-etal-2019-structural} for citation intent classification from NLP, CS, and biomedical articles. As well as the field of research classification using the Microsoft Academic Graph's Paper Field annotations \cite{sinha2015overview}, spanning over seven academic fields. 

\paragraph{Evaluation Metrics} We compute F1 scores for NER (macro average), PICO, REL, and CLS (macro and micro average). To accommodate for the benchmark annotations, NER results are measured on the span-level, PICO on the token-level, while REL and CLS are both reported on the sentence-level. For DEP, we report both the unlabeled attachment score (UAS) and the labeled attachment score (LAS). The scores are averaged over three runs with three random seeds.

\section{Results and Discussion}

Before discussing the results, it is important to note that we fine-tuned all baseline models, SciBERT \cite{beltagy2019scibert}, BERT \cite{devlin-etal-2019-bert}, and RoBERTa \cite{liu2019roberta}, on each task using the optimal hyperparameters provided in their respective publications. We evaluated our two variants \modelNameCustomTokenizer and \modelNameRobertaTokenizer. Table~\ref{tab:benchmarking-results} presents the results across the five task categories and the eleven datasets.

\begin{table*}[t]
    \centering
    \small
    \setlength{\tabcolsep}{2.5pt}
    \begin{tabular}{ccrrrrrr}
        \toprule
         & \multirow{2}{*}{\textbf{Dataset}} & \multicolumn{2}{c}{\textbf{\generalModelName (ours)}} & \multicolumn{1}{c}{\multirow{2}{*}{\textbf{SciBERT}}} & \multicolumn{1}{c}{\multirow{2}{*}{\textbf{BERT}}} & \multicolumn{1}{c}{\multirow{2}{*}{\textbf{RoBERTa}}} & \multicolumn{1}{c}{\multirow{2}{*}{\textbf{ModernBERT}}} \\
         & & \multicolumn{1}{c}{\textbf{Custom}} & \multicolumn{1}{c}{\textbf{RoBERTa}} & & & & \\
        \midrule
        \multirow{4}{*}{\textbf{NER}} & \textbf{BC5CDR}$^{\ddagger}$ & \textbf{97.20±0.10} & 96.54±0.33 & 95.07±1.06 & 85.83±0.17 & 88.07±2.03 & 94.97±0.24 \\
         & \textbf{JNLPBA}$^{\ddagger}$ & 93.89±0.28 & 93.50±0.15 & \textbf{94.19±0.47} & 92.21±1.30 & 92.11±0.47 & 91.51±0.31 \\
         & \textbf{NCBI-disease}$^{\ddagger}$ & 91.48±0.21 & \textbf{91.63±0.07} & 87.09±3.20 & 73.65±9.05 & 88.81±0.67 & 87.74±0.83 \\
         & \textbf{SciERC}$^{\ddagger}$ & 41.03±0.45 & 42.49±1.33 & \textbf{54.39±5.74} & 18.12±7.89 & 14.28±2.58 & 22.41±1.17 \\
        \midrule
        \textbf{PICO} & \textbf{EBM-NLP}$^{\ddagger}$ & 77.66±0.10 & \textbf{78.47±0.10} & 78.14±0.74 & 73.78±1.65 & 77.31±0.01 & 73.91±0.74 \\
        \midrule
        \multirow{2}{*}{\textbf{REL}} & \textbf{ChemProt}$^{\dagger}$ & \textbf{83.83±0.55} & 82.64±0.32 & 82.83±1.76 & 75.04±5.75 & 79.52±0.37 & 70.99±1.38 \\
         & \textbf{SciERC}$^{\ddagger}$ & \textbf{81.23±1.61} & 72.10±14.1 & 77.10±5.74 & 56.58±17.6 & 68.23±2.85 & 73.14±1.00 \\
        \midrule
        \multirow{3}{*}{\textbf{CLS}} & \textbf{CitationIntent}$^{\ddagger}$ & \textbf{64.46±0.46} & 59.08±3.52 & 47.33±15.3 & 29.58±6.78 & 44.37±3.50 & 50.30±6.81 \\
         & \textbf{MAG}$^{\ddagger}$ & 73.01±0.09 & 73.59±0.05 & \textbf{74.61±0.29} & 74.12±0.65 & 74.23±0.09 & 72.27±0.05 \\
         & \textbf{SciCite}$^{\ddagger}$ & 83.96±0.22 & 80.75±6.24 & \textbf{85.49±0.63} & 83.91±0.10 & 84.45±0.15 & 84.22±0.18 \\
        \midrule
        \multirow{2}{*}{\textbf{DEP}} & \textbf{GENIA (LAS)}$^{\P}$ & \textbf{53.33±0.72} & 53.23±1.88 & 36.31±8.55 & 24.15±9.25 & 34.15±2.80 & 30.08±0.97 \\
         & \textbf{GENIA (UAS)}$^{\S}$ & 57.88±0.75 & \textbf{58.57±2.22} & 41.68±8.60 & 26.62±4.52 & 40.44±2.93 & 34.55±1.33 \\
        \midrule 
        \multicolumn{2}{c}{\textbf{Average}} & \textbf{74.91±0.14} & 73.55±1.77 & 71.19±3.97 & 59.47±4.94 & 65.50±0.99 & 65.51±0.55 \\
        \bottomrule
    \end{tabular}
    \caption{Performance comparison (\%) of our models (\generalModelName) and existing language models on five tasks and eleven datasets. Our \modelNameCustomTokenizer and \modelNameRobertaTokenizer models are equipped with a tokenizer trained on our dataset and the tokenizer from existing RoBERTa-base model, respectively. The scores are the F1 scores from each task, averaged over three runs. Notation: $^{\dagger}$ micro-averaged F1 score;  $^{\ddagger}$ macro-averaged F1 score; $^{\P}$ labeled attachment score; $^{\S}$ unlabeled attachment score.}
    \label{tab:benchmarking-results}
\end{table*}

Across all evaluated tasks, our models perform comparably to established domain-specific models, with an overall average score of \textbf{74.91} for \modelNameCustomTokenizer and \textbf{73.55} for \modelNameRobertaTokenizer, slightly improving on the performance of SciBERT (\textbf{71.19}). This demonstrates that our dataset enables robust representation learning on scientific text, even when trained exclusively on open-access data.

\paragraph{Named Entity Recognition (NER).}  
On the NER tasks, \modelNameCustomTokenizer achieves the highest F1 score on the BC5CDR dataset (\textbf{97.20}), surpassing all baselines,  performing on par with SciBERT on JNLPBA, and with \modelNameRobertaTokenizer on NCBI-disease. Although performance drops on the SciERC dataset -- where SciBERT remains the strongest -- our models still outperform general-domain baselines such as BERT and RoBERTa, indicating that domain-specific data and tokenization significantly benefit biomedical and scientific entity recognition.

\paragraph{PICO Extraction.}  
For the EBM-NLP dataset, our \modelNameRobertaTokenizer model achieves the best overall performance (\textbf{78.47}), slightly outperforming SciBERT (\textbf{78.14}) and demonstrating that our corpus provides competitive representations for structured medical text extraction.

\paragraph{Relation Extraction (REL).}  
In the relation extraction tasks, \modelNameCustomTokenizer achieves the best score on both ChemProt (\textbf{83.83}) and SciERC (\textbf{81.24}), surpassing all other baselines. This suggests that our scientific corpus captures relational semantics effectively, even across heterogeneous domains.

\paragraph{Text Classification (CLS).}  
In classification tasks such as CitationIntent, MAG, and SciCite, our models perform competitively. 
\modelNameCustomTokenizer obtains the best results on CitationIntent (\textbf{64.46}), outperforming all baselines by a notable margin, while performance on MAG and SciCite remains close to that of SciBERT and RoBERTa. These findings indicate that our models generalize well across citation and intent classification tasks.

\paragraph{Dependency Parsing (DEP).}  
For dependency parsing on GENIA, evaluation metrics are UAS and LAS. UAS does not consider the semantic relation (e.\,g., Subj) used to label the attachment between the head and child, while LAS requires a correct label for each attachment. Both of our variants outperform all baselines, highlighting the potential of our dataset to enhance syntactic understanding in scientific text.

\paragraph{Overall Analysis.}  
In aggregate, the \modelNameCustomTokenizer and \modelNameRobertaTokenizer models perform on par with, or better than, the baselines. These results validate the strength of our dataset and demonstrate that open-access scientific corpora can yield competitive language models without relying on proprietary, restricted, or even illegally acquired data sources. Our findings underline that reproducibility and openness do not come at the expense of performance, and that the proposed dataset provides a strong foundation for advancing research in scientific language processing.

\section{Related Work}

\citet{hu2025surveyscientificlargelanguage} published a comprehensive survey on current LLMs and datasets for the scientific domain, covering comparable datasets and LLMs, which we discuss in this section. Various transformer-based LLMs have been trained on curated scientific corpora, yielding exceptional results on their domain-specific benchmarks \cite{beltagy2019scibert,phan2021scifivetexttotexttransformermodel,taylor2022galacticalargelanguagemodel}. These models are trained using scientific data from online databases such as arXiv\footnote{\label{fn:arxiv}\url{https://arxiv.org}}, PMC\footnote{\label{fn:pmc}\url{https://pmc.ncbi.nlm.nih.gov/tools/openftlist/}}, ACL Anthology\footnote{\label{fn:acl}\url{https://aclanthology.org}}, or Semantic Scholar\footnote{\label{fn:semantic}\url{https://www.semanticscholar.org}}. However, to the best of our knowledge, the articles from Unpaywall have not been used for LLM pre-training to date. For instance, the encoder-based model SciBERT has been trained on a set of papers from Semantic Scholar \cite{ammar2018constructionliteraturegraphsemantic} and evaluated on downstream tasks like NER, PICO, CLS, REL, and DEP \cite{beltagy2019scibert}. The sequence-to-sequence model SciFive \cite{phan2021scifivetexttotexttransformermodel} has been trained on a collection of PMC Abstracts and full texts\footnote{\label{fn:pmc_2}\url{https://pubmed.ncbi.nlm.nih.gov}}.
Larger and more recent models, e.\,g., GALACTICA \cite{taylor2022galacticalargelanguagemodel}, use a collection of multiple corpora from arXiv, PMC, ACL Anthology, and Semantic Scholar. They claim to use a corpus with approximately 88 billion tokens. Note that \emph{none} of the data splits used for training these scientific (L)LMs are publicly available, which makes it impossible to analyze the details of the data coverage.

Based on the previously referenced databases for scientific research, there exist four recent and extensive corpora. The largest publicly available corpus of scientific text is S2ORC \cite{lo2020s2orc}, spanning 81.1 million open-access papers from multiple academic fields and curated from various digital archives. S2ORC is constructed from the PDF files of the Semantic Scholar literature corpus, processed using Grobid, and assembled using the metadata provided by Grobid, for instance, by checking the DOI against other open-access databases, such as Unpaywall \cite{chawla2017unpaywall}. UnarXive \cite{Saier_2023} is a dataset based on publications uploaded to arXiv, reaching over 1.9 million documents. UnarXive also covers multiple scientific disciplines and is not limited to a single scientific field. The ACL Anthology Network (AAN) \cite{radev-etal-2009-acl} is a manually curated, networked database of documents built upon the ACL Anthology. ANN provides a bibliometric-enhanced corpus covering 24.6k papers from computational linguistics and NLP. The fourth corpus is the PMC OA Subset, a dataset comprising journal articles and preprints from PMC. It is important to note that not all articles within PMC are eligible for text mining or other forms of reuse, as a significant proportion remains protected under copyright restrictions. Our contribution extends the current set of public corpora with new datapoints that have not been used for LLM pretraining so far.

\section{Conclusion}
\label{sec:conlusion}

We present \datasetname{}, a new dataset of scientific publications, constructed entirely using open-source frameworks and publicly accessible data sources. The dataset, together with a transparent and reproducible pipeline, demonstrates that large-scale scientific corpora can be developed without reliance on proprietary infrastructure or restricted resources.
By harvesting open-access texts from Unpaywall, arXiv, and PLOS, our processing and normalization workflow establishes a scalable, ethical blueprint for data creation in natural scientific language processing (NSLP).

The way the Unpaywall OA list is exploited, and the modular approach in the pipeline, allow SciLaD to be incrementally maintained and extended with new scientific articles without rerunning the entire harvesting and preprocessing pipeline from scratch. 

Beyond the dataset, we introduce an extensible evaluation pipeline that enables fair and consistent benchmarking across diverse model architectures, including encoder-only, decoder-only, and sequence-to-sequence models. By making both the dataset and the evaluation framework openly available, we aim to support the broader community in advancing domain-specific NLP research. This contribution underscores the importance of transparency, accessibility, and reproducibility as guiding principles for future scientific language technologies.


\section*{Limitations}
\label{sec:limitations}

This work has several limitations that should be addressed in future releases of the dataset. 

\paragraph{Open Source Data Split}  
While the dataset presented in this work is fully derived from open-access resources, the current release focuses primarily on the textual content and associated metadata. In future iterations, we aim to provide a clearly defined \textsc{CC-BY} data split that ensures unrestricted use and redistribution across academic and industrial settings. This release will also include additional per-sample information, such as corresponding PDF file versions of abstracts, enabling richer downstream tasks such as layout analysis and document structure modeling. Moreover, due to language distribution imbalance, the \cleanPlainDataset in this study is limited to English, and expanding it to include other languages would enhance its applicability across diverse linguistic contexts. 

\paragraph{Data Curation}  
Although our dataset construction pipeline captures a wide variety of scientific publications, there remains room for improvement in terms of fine-grained curation. In particular, future work will extend entity and relation extraction, as well as structured table and formula parsing, to support more complex information retrieval and reasoning tasks. These improvements are crucial for developing high-quality datasets that can facilitate scientific table-to-text generation and enhance the training of specialized scientific LLMs.

\paragraph{LLM Training}  
Our current experiments focus on discriminative tasks such as classification, relation extraction, and named entity recognition. However, the dataset also holds potential for generative model development. As part of our ongoing work, we plan to train sequence-to-sequence and decoder-only language models—such as compact architectures in the style of \textit{SmolLM2} \cite{allal2025smollm2smolgoesbig} to explore the dataset’s applicability for summarization, text generation, and scientific question answering. This direction will enable a broader evaluation of how open, domain-specific corpora can support scalable and transparent LLM training.

\paragraph{Citation graph exploitation} 
The data are represented in TEI XML, which preserves the document structure and citation graphs. Those TEI files contain valuable information that can be used for pretraining LLMs or for fine-tuning of downstream tasks, for instance. However, in our project, we pre-trained the LLMs only on plain text. Moreover, we evaluated our model on 12 downstream biomedical tasks. Those benchmarks evaluate on major NLP tasks in the scientific domain, such as NER, REL, and CLS. Although we believe the results demonstrate improvements in scientific language modeling with our dataset, the focus could be on more recent tasks, such as long-context reasoning, full-document summarization, and sophisticated citation graph modeling. Training a language model on TEI XML would most likely also improve performance on these tasks. Due to computational constraints, we aim to use the data for a future project to train a more sophisticated LLM that will then be evaluated on more recent downstream tasks.
In future work, we envision a targeted exploitation of the citation graph as well as keyword mapping, which would help to assess and improve the quality of the data in the following releases. 

\section*{Data and code availability}
\label{sec:availability}

The data presented in this article consists of 35.3 million documents structured in TEI XML, which is limited to research activities (\url{https://huggingface.co/datasets/scilons/SciLaD-all-xml-v1}). 
The full text-based version of the dataset is available at \url{https://huggingface.co/datasets/scilons/SciLaD-all-text-v1}
The English deduplicated version is available as parquet at \url{https://huggingface.co/datasets/scilons/SciLaD-en-dedup-v1}. 
A split consisting of only Creative Commons licenses that allow redistribution (see above) will be shared separately without limitations in a future release.

The code is available on GitHub. The harvesting project is available at \url{https://github.com/kermitt2/article_dataset_builder}. 
The evaluation pipeline is hosted at \url{https://github.com/scilons/scilons-eval}, the processing pipeline is available at \url{https://github.com/scilons/harvesting}, the training code and scripts for \modelNameCustomTokenizer and \modelNameRobertaTokenizer are available at \url{https://github.com/scilons/roberta-pretrain}. The fork of \LaTeX{ML} used in this work is available at \url{https://github.com/kermitt2/LaTeXML/}.
The rest of the tools used in this work, such as Grobid~\cite{Grobid}, Pub2TEI~\cite{pub2tei} are available in the cited repository.

The pre-trained models are available on Huggingface, the \modelNameCustomTokenizer final version is at \url{https://huggingface.co/scilons/SciLaD-M-custom} and \modelNameRobertaTokenizer is at \url{https://huggingface.co/scilons/SciLaD-M-roberta}.

All checkpoints of \modelNameCustomTokenizer are available at \url{https://huggingface.co/collections/scilons/scilad-m-custom}, the custom tokenizer is available at \url{(https://huggingface.co/scilons/sciroberta-tokenizer-v3}. 
All checkpoints of \modelNameRobertaTokenizer are available at \url{https://huggingface.co/collections/scilons/scilad-m-roberta}.

\section{Acknowledgments}


The data collection was provided by James Howison and Patrice Lopez in the scope of the SoftCite project\footnote{\url{https://github.com/softcite}}, aiming to create a large dataset of ML-identified mentions of software~\cite{howison_2025_15149379}.
The harvesting pipeline was run through the JetStream 2 program at Indiana University\footnote{\url{https://jetstream-cloud.org/}} ran through allocation CIS220172 from the Advanced Cyberinfrastructure Coordination Ecosystem: Services \& Support (ACCESS) program, which is supported by National Science Foundation grants \#2138259, \#2138286, \#2138307, \#2137603, and \#2138296. 
The authors also acknowledge the Texas Advanced Computing Center (TACC)\footnote{\url{http://www.tacc.utexas.edu}} at The University of Texas at Austin for providing computational resources that have contributed to the creation and processing of this research dataset.

This work was supported by the consortium NFDI for Data Science and Artificial Intelligence (NFDI4DS)\footnote{\url{https://www.nfdi4datascience.de}} as part of the non-profit association National Research Data Infrastructure (NFDI e.\,V.). The consortium is funded by the Federal Republic of Germany and its states through the German Research Foundation (DFG) project NFDI4DS (no.~460234259).
Further support was provided from the European Union’s Horizon Europe research and innovation programme under grant agreement No.~101189745 (HIVEMIND).

Finally, the authors acknowledge the University of Mannheim and the Deutsches Forschungszentrum für Künstliche Intelligenz GmbH (DFKI) for providing the necessary resources required for training and evaluating the models. 




\section*{Bibliographical References}
\label{sec:reference}

\bibliographystyle{lrec2026-natbib}
\bibliography{custom}

\clearpage

\begin{minipage}{\textwidth}
\centering
\captionsetup{type=figure}

\begin{subfigure}[c]{0.53\textwidth}
  \centering
  \includegraphics[width=\textwidth]{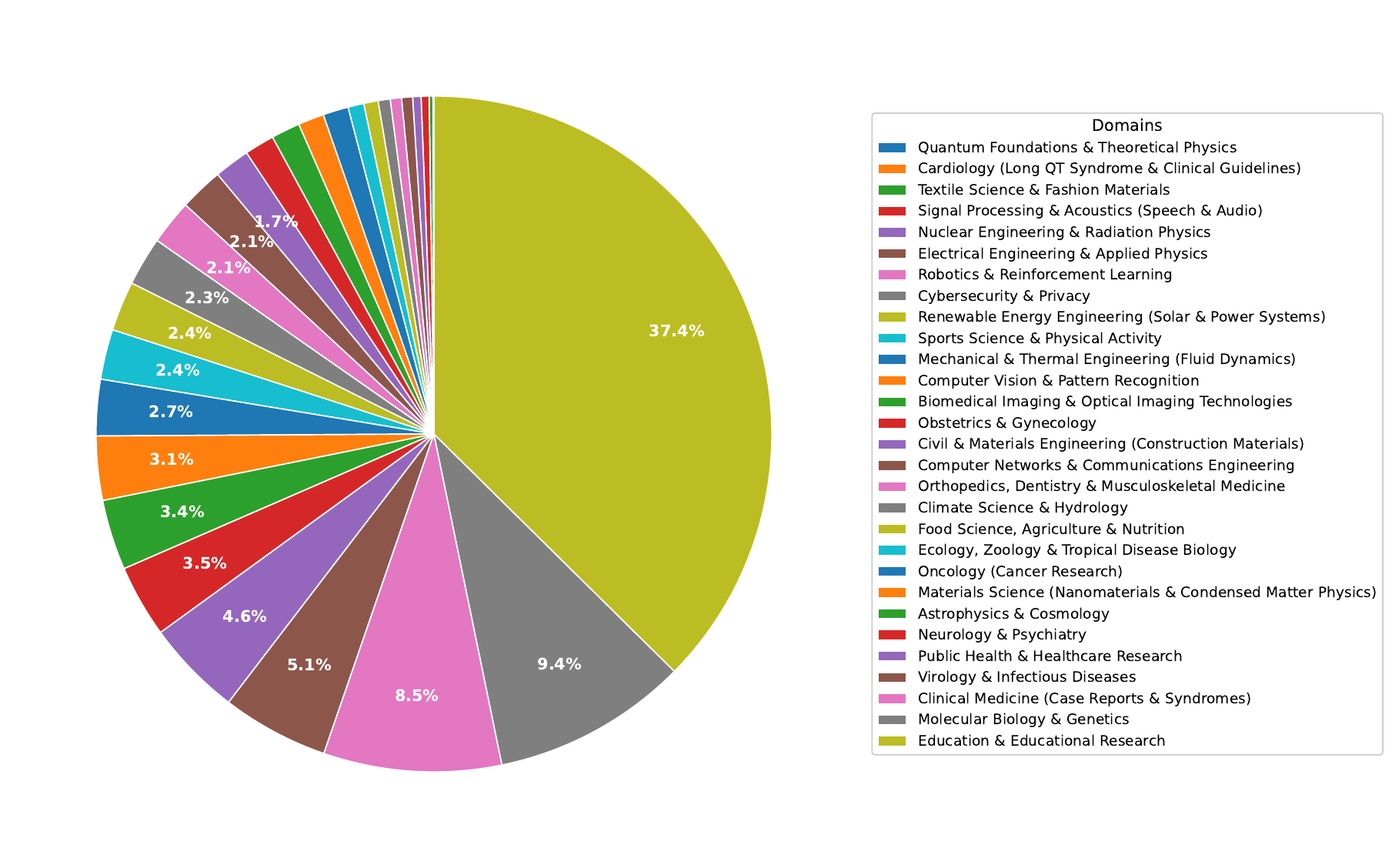}
  \caption{Fine-grained domain distribution}
  \label{sub:fig:domains-fine}
\end{subfigure}
\hfill
\begin{subfigure}[c]{0.45\textwidth}
  \centering
  \includegraphics[width=\textwidth]{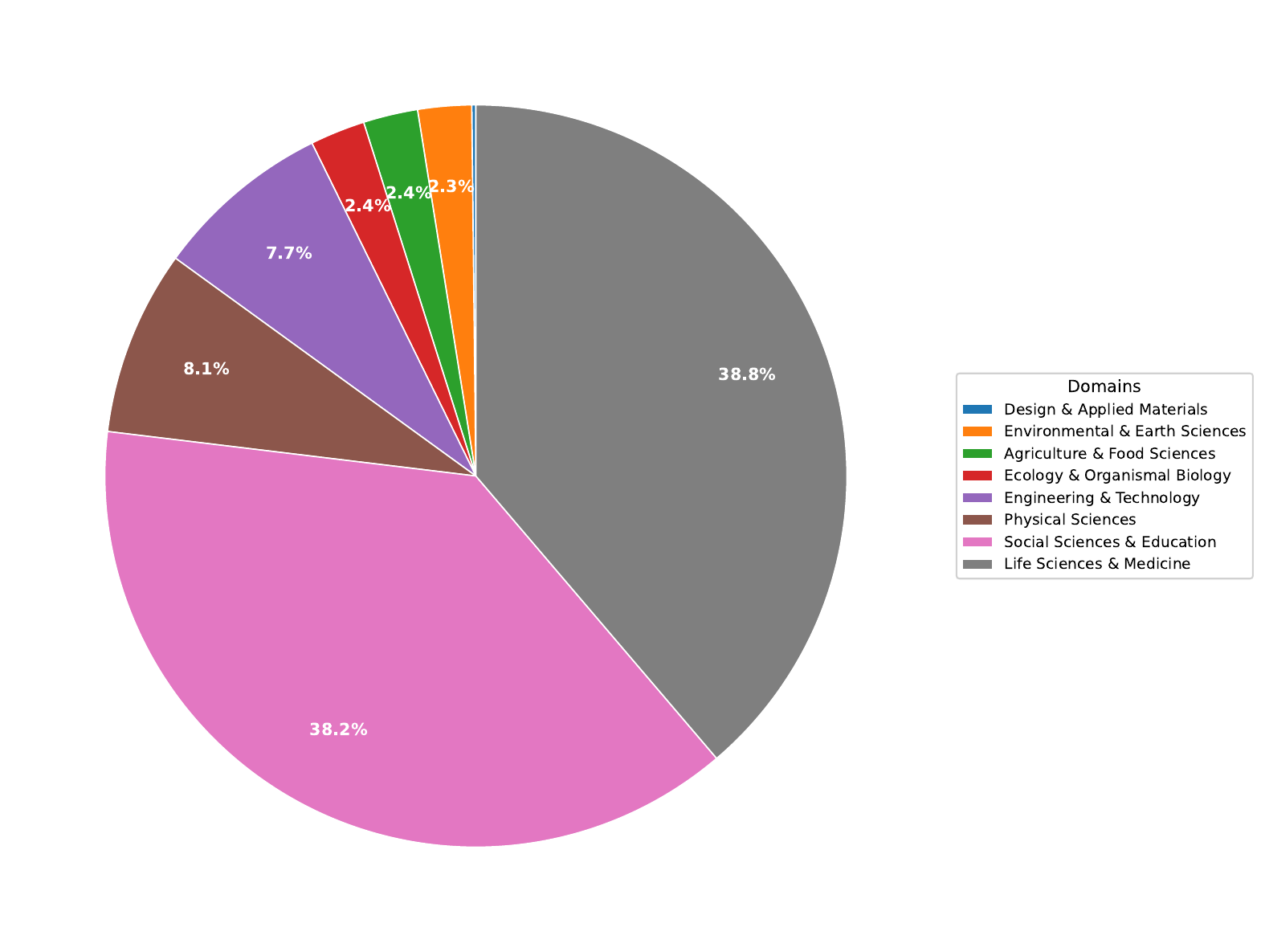}
  \caption{High-level domain distribution}
  \label{sub:fig:domains-high}
\end{subfigure}

\caption{Distribution of domains in the full dataset.}
\label{fig:domains}
\end{minipage}

\appendix
\section{Domain Distribution}
\label{sec:app-domains}

To showcase the available scientific domains in the dataset, we apply a two-stage topic modeling approach. First, a representative sample of one million paper titles is drawn from the dataset using reservoir sampling, ensuring a uniform sample of the corpus. Each title was then encoded using SentenceTransformers\footnote{Specifically: \url{https://huggingface.co/sentence-transformers/all-MiniLM-L6-v2}}, and all embeddings were given to BERTopic~\cite{grootendorst2022bertopic} to produce 30 fine-grained topic clusters. We then labeled the resulting topics using GPT5~\cite{singh2025openai} based on the keywords and representative document produced by BERTopic. The topics were manually reviewed and clustered further into 8 high-level scientific domains. Once the topic model was trained, we applied a second straming pass across the full corpus, assigning each title to the closest cluster using the same embedding method. Figure~\ref{sub:fig:domains-fine} shows the fine-grained topic distribution and Figure~\ref{sub:fig:domains-high} the high-level domains in \datasetname{}.

\end{document}